\documentclass{elsart}
\usepackage{amssymb}
\usepackage{subfigure}
\usepackage{lineno}
\usepackage{booktabs}
\usepackage{threeparttable}
\usepackage[toc,page,title,titletoc,header]{appendix}

\usepackage{graphicx}
\usepackage{verbatim}
\usepackage{amsmath}
\usepackage{longtable}
\usepackage[T1]{fontenc}
\usepackage[latin1]{inputenc}
\usepackage{cases}
\usepackage[usenames]{color}
\usepackage{algorithmic}
\usepackage{algorithm}
\usepackage{rotating}
\usepackage{url}

\begin{document}
%%\begin{linenumbers}
\begin{frontmatter}



\title{On memetic search for the max-mean dispersion problem}


\author{Xiangjing Lai}
\ead{laixiangjing@gmail.com}
\and
\author{Jin-Kao Hao\corauthref{cor}}
\ead{hao@info.univ-angers.fr}
\corauth[cor]{Corresponding author.}
%%\cortext[cor]{Corresponding author.}
\address{LERIA, Universit$\acute{e}$ d'Angers, 2 Bd Lavoisier, 49045 Angers, Cedex 01, France \\
\textbf{Submitted for publication on 19 January 2015}}

\begin{abstract}
Given a set $V$ of $n$ elements and a distance matrix $[d_{ij}]_{n\times n}$ among elements, the max-mean dispersion problem (MaxMeanDP) consists in selecting a subset $M$ from $V$ such that the mean dispersion (or distance) among the selected elements is maximized. Being a useful model to formulate several relevant applications, MaxMeanDP is known to be NP-hard and thus computationally difficult. In this paper, we present a highly effective memetic algorithm for MaxMeanDP which relies on solution recombination and local optimization to find high quality solutions. Computational experiments on the set of 160 benchmark instances with up to 1000 elements commonly used in the literature show that the proposed algorithm improves or matches the published best known results for all instances in a short computing time, with only one exception, while achieving a high success rate of 100\%. In particular, we improve 59 previous best results out of the 60 most challenging instances. Results on a set of 40 new large instances with 3000 and 5000 elements are also presented. The key ingredients of the proposed algorithm are investigated to shed light on how they affect the performance of the algorithm.

\end{abstract}

\end{frontmatter}
%\linenumbers

\section{Introduction}
\label{Intro}

Given a weighted complete graph $G = (V, E, D)$, where $V$ is the set of $n$ vertices, $E$ is the set of $\frac{n\times(n-1)}{2}$ edges, and $D$ represents the set of edge weights $d_{ij}$ ($i\neq j$), the generic equitable dispersion problem consists in selecting a subset $M$ from $V$ such that some objective function $f$ defined on the subgraph induced by $M$ is optimized \cite{Prokopyev2009}. In the related literature, a vertex $v \in V$ is also called an element, and the edge weight $d_{ij} \in D$ is called the distance (or diversity) between elements $i$ and $j$.

According to the objective function to be optimized as well as the constraints on the cardinality of subset $M$, several specific equitable dispersion problem can be defined. At first, if the cardinality of $M$ is fixed to a given number $m$, the related equitable dispersion problems include the following four classic variants: (1) the max-sum diversity problem, also known as the maximum diversity problem (MDP), which is to maximize the sum of distances among the selected elements \cite{Aringhieri2008,Aringhieri2011,Duarte2007,Glover1998,Marti2010,Palubeckis2007,Wu2013b}; (2) the max-min diversity problem that aims to maximize the minimum distance among the selected elements \cite{DellaCroce2009,Porumbel2011,Resendea2010,Saboonchi2014}; (3) the maximum minsum dispersion problem (MaxMinsumDP) that aims to maximize the minimum aggregate dispersion among the selected elements \cite{Aringhieri2014,Prokopyev2009}; (4) the minimum differential dispersion problem (MinDiffDP) whose goal is to minimize the difference between the maximum and minimum aggregate dispersion among the selected elements to guarantee that each selected element has the approximately same total distance from the other selected elements \cite{Aringhieri2014,Duarte2014,Prokopyev2009}. In addition, when the cardinality of subset $M$ is not fixed, i.e., the size of $M$ is allowed to vary from 2 to $n$, the related equitable dispersion problems include the max-mean dispersion problem (MaxMeanDP) and the weighted MaxMeanDP \cite{Carrasco2014,DellaCroce2014,Marti2013,Prokopyev2009}.

In this study, we focus on MaxMeanDP which can be described as follows \cite{Prokopyev2009}. Given a set $V$ of $n$ elements and a distance matrix $(d_{ij})_{n \times n}$ where $d_{ij}$ represents the distance between elements $i$ and $j$ and can take a positive or negative value, the max-mean dispersion problem consists in selecting a subset $M$ ($|M|$ is not fixed) from $V$ such that the mean dispersion among the selected elements, i.e., $\frac{\sum_{i,j\in M;i<j} d_{ij}}{|M|} $, is maximized.

MaxMeanDP can be naturally expressed as a quadratic integer program with binary variables $x_i$ that takes 1 if element $i$ is selected and 0 otherwise \cite{Marti2013,Prokopyev2009}, i.e.,
\begin{equation}\label{FMAX}
\mathrm{Maximize} \quad f(s) = \frac{\sum_{i=1}^{i=n-1} \sum_{j=i+1}^{n} d_{ij} x_i x_j} {\sum_{i=1}^{n} x_i}
\end{equation}
\begin{equation}\label{size constraint}
\mathrm{Subject \ to} \quad \sum_{i=1}^{n} x_{i} \ge 2
\end{equation}
\begin{equation} \label{binary constraint}
\quad x_{i}\in \{0,1\},  i=1,2,\dots, n;
\end{equation}
where the constraint (\ref{size constraint}) guarantees that at least two elements are selected.

In addition to its theoretical signification as a strongly NP-hard problem \cite{Prokopyev2009}, MaxMeanDP is notable for its ability to model a variety of real-world applications, such as web pages ranks \cite{Kerchove}, community mining \cite{Yang}, and others mentioned in \cite{Carrasco2014}.

Given the interest of MaxMeanDP, several useful solution approaches have been proposed in the literature to deal with this hard combinatorial optimization problem. First of all, Prokopyev et al. presented a mixed-integer 0-1 linear programming formulation and solved small instances with up to 100 elements with the CPLEX solver \cite{Prokopyev2009}. In the same work, they also introduced a GRASP method and made a comparison between the GRASP method and the MILP method to show the superiority of GRASP. Subsequently, Mart\'{i} and Sandoya proposed a GRASP with the path relinking method (GRASP-PR) \cite{Marti2013}, and the computational results show that GRASP-PR outperforms the previously reported methods. Very recently, Della et al. reported a hybrid heuristic approach based on a quadratic knapsack formulation \cite{DellaCroce2014}, and their computational experiment shows that the hybrid approach is superior to the GRASP-PR method. In another very recent work, Carrasco et al. proposed a diversified tabu search algorithm by combining a short-term tabu search procedure and a long-term tabu search procedure \cite{Carrasco2014}, and the computational results show that this algorithm clearly dominates the previous GRASP-PR method.

In this paper, we propose the first population-based memetic algorithm for solving MaxMeanDP (called MAMMDP). The proposed algorithm combines a random crossover operator to generate new offspring solutions and a tabu search method to find good local optima.

The performance of our algorithm is assessed on a set of 160 benchmark instances ($20\le n \le 1000$) commonly used in the literature and a set of additional 40 large-sized instances that we generate ($n=3000,5000$). For the first set of existing benchmarks, the experimental results show that the proposed algorithm is able to attain, in a short or very short computing time, all current best known results established by any existing algorithms, except for one instance. Furthermore, it can even improve the previous best known result for a number of these instances. The effectiveness of the proposed algorithm is also verified on much larger instances of the second set with 3000 and 5000 elements.

The remaining part of the paper is organized as follows. In Section \ref{method}, we describe in detail the general scheme and the ingredients of the proposed algorithm. In Section \ref{results}, we present the computational results based on the 200 benchmark instances and compare them with those of the existing state-of-the-art algorithms from the literature. In Section \ref{Discussion}, some important ingredients of the proposed algorithm are analyzed and discussed. Finally, we conclude the paper in the last Section.

\section{Memetic Algorithm for Max-Mean Dispersion Problem}
\label{method}

Memetic search is a well-known metaheuristic framework which aims to provide the search with a desirable trade-off between intensification and diversification through the combined use of a crossover operator (to generate new promising solutions) and a local optimization procedure (to locally improve the generated solutions) \cite{Moscato2003,Neri2011}. %Memetic search has been applied with success to a number of combinatorial problems like bin packing \cite{Falkenauer}, graph coloring \cite{Galinier1999,Lu,Porumbeletal2010}, graph partition \cite{BenlicHao2011,Galinier2011,MerzFreisleben2000,Wu2013a}, and maximum diversity \cite{Wu2013b}. %, and vehicle routing \cite{Matei2015} bicluster identification of DNA microarray data \cite{Ayadi},

\subsection{General Procedure}

The proposed memetic algorithm (denoted by MAMMDP) adopts the principles and guidelines of designing effective MA for discrete combinatorial problems \cite{Hao2012}. Indeed, the overall performance of a memetic algorithm depends largely on the implementation of these two key components that need to be carefully designed according to the specific problem structure. Additionally, the proposed MAMMDP algorithm uses a population scheme which borrows from the path relinking method \cite{Glover2000} to ensure a strong intensification search.

The general procedure of our MAMMDP algorithm is shown in Algorithm \ref{Algo_MA_MaxMeanDP}, where $s^*$ and $s^w$ respectively represent the best solution found so far and the worst solution in the population in terms of the objective value, and $PairSet$ is the set of solution pairs $(s^i, s^j)$, which is initially composed of all the possible solution pairs $(s^i, s^j)$ in the population and is dynamically updated as the search progresses.

Our MAMMDP algorithm starts with an initial population $P$ (line 4) which includes $p$ different solutions, where each of them is randomly generated and then improved by the tabu search procedure. After the initialization of population (Section \ref{subsec_Initialization_population}), the algorithm enters a while loop (lines 11 to 25) to make a number of generations. At each generation, a solution pair $(s^i,s^j)$ is randomly selected from $PairSet$ and then the crossover operator (line 14) is applied to the selected solution pair $(s^i,s^j)$ to generate a new solution $s^{o}$ (Section \ref{subset_crossover}). Subsequently, $s^{o}$ is improved by the tabu search procedure (line 15) (Section \ref{subsec_LS}). After that, a population updating rule is used to update the population (lines 20 to 24) (Section \ref{subsec_pool_updating}). Meanwhile, the $PairSet$ is accordingly updated as follows: First, the solution pair $(s^i, s^j)$ is removed from $PairSet$ (line 13); Then, if an offspring solution $s^{o}$ replaces the worst solution $s^w$ in the population, all the solution pairs containing $s^w$ are removed from $PairSet$ and all the solution pairs that can be generated by combining $s^{o}$ with other solutions in the population are added into $PairSet$ (lines 23 to 24). The while loop ends when $PairSet$ becomes empty, then the population is recreated, while preserving the best solution ($s^{*}$) found so for in the new population (lines 4 to 8), and the above while loop is repeated if the timeout limit is not reached.

It is worth noting that compared with the traditional random selection scheme, the proposed MAMMDP algorithm uses the set $PairSet$ to contain the solution pairs of the population for crossover operations. This strategy ensures that every pair of solutions in the population is combined exactly once, favoring an more intensified search.

\begin{algorithm}[!htbp]
\begin{small}
 \caption{Memetic algorithm for Max-mean Dispersion Problem} \label{Algo_MA_MaxMeanDP}
 \begin{algorithmic}[1]
   \STATE \sf \textbf{Input}: The set $V=\{v_1,v_2,\dots,v_n\}$ of $n$ elements and the distance matrix $D=[d_{ij}]_{n \times n}$, the population size $p$, the timeout limit $t_{out}$.
   \STATE \textbf{Output}: the best solution $s^{*}$ found% and its objective value $f(s^{*})$
   \REPEAT
   \STATE $P=\{s^{1},\ldots,s^{p}\}$ $\leftarrow$
   Population\_Initialization($V$) \ \ \ \ \ \ \ \ \ \ \ \ \ \ \ /$*$ Section \ref{subsec_Initialization_population} $*$/
  % \STATE $s^{*} \leftarrow Best(Pop)$ \ \ \ \  /$*$ $s^{*}$ keeps the best solution found by the algorithm  $*$/
   \IF{it is not in the first loop}
   \STATE $s^{w} \leftarrow arg \ min \{f(s^{i}) : i=1,\ldots,p\}$
   \STATE $P \leftarrow P \cup \{s^{*}\} \setminus \{s^{w}\}$
   \ENDIF
   \STATE $s^{*} \leftarrow arg \ max \{f(s^{i}) : i=1,\ldots,p\}$ \ \ \ \ \ \ \ /$*$ $s^{*}$ keeps the best solution found $*$/
   \STATE $PairSet \leftarrow \{(s^i,s^j) : 1\leq i < j \leq p \}$
   \WHILE{$PairSet\neq \emptyset$ and $time < t_{out}$}
        \STATE Randomly pick a solution pair $(s^i,s^j)\in PairSet$
        \STATE $PairSet \leftarrow PairSet \setminus \{(s^i,s^j)\}$
        \STATE $s^{o} \leftarrow Crossover Operator(s^i,s^j)$ \ \ \ \ \ \ \ \ \ \ \ \ \ \ \ \ \ \ \ \ \ \ \ \ \ \ \ \ \ \ \ \ \ /$*$ Section \ref{subset_crossover} $*$/
        \STATE $s^{o}$ $\leftarrow$ $Tabu Search(s^{o})$  \ \ \ \ \ \ \ \ \ \ \ \ \ \ \ \ \ \ \ \ \ \ \ \ \ \ \ \ \ \ \ \ \ \ \ \ \ \ \ \ \ \ \ \ \ \ \ /$*$ Section \ref{subsec_LS} $*$/
        \IF{$f(s^{o}) > f(s^{*})$}
              \STATE $s^{*} \leftarrow s^{o}$
        \ENDIF
        \STATE $s^{w} \leftarrow arg \ min \{f(s^{i}) : i=1,\ldots,p\}$
        \IF{$s^{o}$ dose not exist in $P$ and $f(s^{o})>f(s^{w})$}
        \STATE $P \leftarrow P \cup \{s^{o}\} \setminus \{s^{w}\}$
        \STATE $PairSet \leftarrow PairSet \setminus \{(s^w,s^k): s^k \in P\}$
        \STATE $PairSet \leftarrow PairSet \cup \{(s^{o},s^k): s^k \in P\}$
        \ENDIF  \ \ \ \ \ \ \ \ \ \ \ \ \ \ \ \ \ \ \ \ \ \ \ \ \ \ \ \ \ \ \ \ \ \ \ \ \ \ \ \ \ \ \ \ \ \ \ \ \ \ \ \ \ \ \ \ \ \ \ \ \ \ \ \ \ \ \ /$*$ Section \ref{subsec_pool_updating} $*$/
   \ENDWHILE
   \UNTIL{$time \ge t_{out}$}
 \end{algorithmic}
 \end{small}
\end{algorithm}

\subsection{Search Space and Solution Representation}
\label{subsec_space}

Given a MaxMeanDP instance denoted by a set $V$ of $n$ elements as well as its distance matrix $D = [d_{ij}]_{n \times n}$, the search space $\Omega$ explored by our MAMMDP algorithm is composed of all possible subsets of $V$, i.e, $\Omega = \{ M: M \subseteq V\}$. Formally, a subset $M$ of $V$ can be expressed by a $n$-dimensional binary vector, $(x_1,x_2,\dots, x_n)$, where $x_i$ takes 1 if element $i$ belongs to $M$, and 0 otherwise. In other words, the search space $\Omega$ is composed of all possible $n$-dimensional binary vectors, i.e.,
\begin{center}
$\Omega = \{(x_1,x_2,\dots, x_n): x_i \in \{0,1\}, 1\le i \le n \}$
\end{center}
Clearly, the size of the search space $\Omega$ is bounded by $O(2^n)$.

For any candidate solution $s=(x_1,x_2,...,x_n) \in \Omega$, its quality is given by the objective value ($f(s)$, Formula (\ref{FMAX})) of the max-mean dispersion problem.

\subsection{Population Initialization}
\label{subsec_Initialization_population}

In our memetic algorithm, the initial population of $p$ solutions is generated as follows. First, we generate $p$ random solutions, where each component $x_i$ ($i=1,2,\dots,n$) of a solution $(x_1,x_2,...,x_n)$ is randomly assigned a value from $\{0,1\}$ using a uniform probability distribution.  Then, the tabu search method (see Section \ref{subsec_LS}) is applied to each of the generated solutions to optimize them to a local optimum solution, and the resulting solutions are used to form the initial population.

\subsection{Local Optimization using Tabu Search}
\label{subsec_LS}
Local optimization is a key component of a memetic algorithm and ensures generally the role of an intensified search to locate high quality local optimum. In this study, we devise a tabu search (TS) method as the local optimization procedure which proves to be highly effective when when it is applied alone.

Given a neighborhood structure ($N(s)$) and a starting solution ($s_0$), our tabu search procedure iteratively replaces the incumbent solution $s$ by a best eligible neighboring solution ($s^{'}$) until the stopping condition is met, i.e., the best solution ($s_b$) is not improved for $\alpha$ consecutive iterations (called the depth of TS). At each iteration of TS, the performed move is recorded in the tabu list to prevent the reverse move from being performed for the next $tt$ iterations. Here, $tt$ is called the tabu tenure and controlled by a special tabu list management strategy. Note that in this tabu search method a move is identified to be eligible if it is not forbidden by the tabu list or it leads to a solution better than the best solution found so far in terms of the objective function value (aspiration criterion).

The general scheme of our TS method is described in Algorithm \ref{Algo_TS}, and the neighborhood structure employed by our TS method and the tabu list management strategy are described in the following subsections.

\begin{algorithm}[!htbp]
\begin{small}
 \caption{$TabuSearch(s_0,N(s),\alpha)$} \label{Algo_TS}
 \begin{algorithmic}[1]
   \STATE \sf \textbf{Input}: Input solution $s_0$, neighborhood $N(s)$, search depth $\alpha$
   \STATE \textbf{Output}: The best solution $s_b$ found during the tabu search process
   \STATE $s \leftarrow s_0$   \quad /* $s$ is the current solution */
	\STATE $s_b \leftarrow s$    \quad /* $s_b$ is the best solution found so far */
	 \STATE $d=0$  \ \quad /* $d$ counts the consecutive iterations where $s_b$ is not updated */
   \REPEAT
   \STATE Choose a best eligible neighboring solution $s' \in N(s)$
   \STATE $s \leftarrow s'$
   \STATE Update tabu list
	 \IF {$f(s) > f(s_b)$}
    \STATE $s_b \leftarrow s$,
		\STATE $d=0$
		\ELSE
		\STATE $d=d+1$
   \ENDIF
   \UNTIL{$d = \alpha$}
	\RETURN $s_b$
 \end{algorithmic}
 \end{small}
\end{algorithm}

\subsubsection{Move and Neighborhood}
\label{TS_NS}

The neighborhood $N_1$ of our tabu search algorithm is defined by the one-flip move operator which consists of changing the value of a single variable $x_i$ to its complementary vale $1 - x_i$.  As such, given a solution $s$, the one-flip neighborhood $N_1(s)$ of $s$ is composed of all possible solutions that can be obtained by applying the one-flip move to $s$. The size of the neighborhood $N_1(s)$ is thus bounded by $O(n)$, where $n$ is the number of components in $s$.

\subsubsection{Fast Neighborhood Evaluation Technique}
\label{Neighborhood_Evaluation}

In our TS method, we employ a fast neighborhood evaluation technique to examine the neighborhood $N_1$. For this purpose, we maintain a $n$-dimensional vector $W=(p_1,p_2,\dots,p_n)$ to effectively calculate the move value (i.e., the change of objective value) of each possible move applicable to the current solution, where the entry $p_i$ represents the sum of distances between the element $i$ and the selected elements for the current solution, i.e., $p_i = \sum_{j\in M; j \neq i} d_{ij}$, where $M$ is the set of selected elements.

Given the current solution $s$, if an one-flip move is performed, i.e., a variable $x_i$ is flipped as $x_i \leftarrow (1-x_i)$, then the move value $\Delta_i$ can be rapidly computed as follows:

\begin{numcases}{\Delta_i = }
\frac{-f(s)}{|M|+1}+\frac{p_i}{|M|+1},& for $x_i=0$;\\
\frac{f(s)}{|M|-1}-\frac{p_i}{|M|-1},& for $x_i=1$;
\end{numcases}
where $f(s)$ is the objective value of the current solution $s$ and $|M|$ is the number of selected elements in $s$. Subsequently, the vector $W$ is accordingly updated as:
\begin{numcases}{p_j=}
p_j + d_{ij},& for $x_i=0,j\neq i$;\\
p_j - d_{ij},& for $x_i=1,j\neq i$;\\
p_j,& for $j=i$;
\end{numcases}
Note that the vector $W$ is initialized at the beginning of each call of TS with the complexity of $O(n^2)$, and is updated in $O(n)$ after each move.

\subsubsection{Tabu List Management Strategy}
\label{TS_LIST}

In our TS procedure, we use a tabu list management strategy to dynamically tune the tabu tenure $tt$, which is adapted according to a technique proposed in \cite{Galinier2011} where the tabu tenure is given by a periodic step function. If the current iteration is $y$, then the tabu tenure of a move is denoted by $tt(y)$.

Precisely, our tabu tenure function is defined, for each period, by a sequence of values $(a_1,a_2,\cdots,a_{q+1})$ and a sequence of interval margins $(y_1,y_2,\cdots,y_{q+1})$ such that for each $y$ in  $[y_i,y_{i+1}-1]$, $tt(y) = a_i+rand(2)$, where $rand(2)$ denotes a random integer between 0 to 2. Here, $q$ is fixed to 15, $(a)_{i=1,\cdots,15}=\frac{T_{max}}{8}(1,2,1,4,1,2,1,8,1,2,1,4,1,2,1)$, where $T_{max}$ is a parameter which represents the maximum tabu tenure. Finally, the interval margins are defined by $y_1=1$, $y_{i+1}=y_i + 5a_i$ ($i\le15$).

Thus, this function varies periodically and for each period, 15 tabu tenures are used dynamically, each being kept for a number of consecutive iterations. In principle, this function helps the tabu search procedure reach a desirable tradeoff between intensification and diversification during its search.

\subsection{Crossover Operator}
\label{subset_crossover}

\begin{algorithm}[!htbp]
\begin{small}
 \caption{The Uniform Crossover Operator for MaxMeanDP} \label{Algo_random_crossover}
 \begin{algorithmic}[1]
   \STATE \sf \textbf{Input}: Two parent solutions $s^1=(x_1^{1},x_2^{1},\dots,x_n^{1})$ and $s^2=(x_1^{2},x_2^{2},\dots,x_n^{2})$.
   \STATE \textbf{Output}:  Offspring solution $s^o=(x_1^{o},x_2^{o},\dots,x_n^{o})$  \\
   \FOR{$i=1$ to $n$}
   \STATE $r$ $\leftarrow$ $rand[0,1)$  \ \ /* rand[0,1) denotes a random number between 0 and 1 */
   \IF{$r < 0.5$}
      \STATE $x_i^{o} \leftarrow x_i^{1}$
   \ELSE
      \STATE $x_i^{o} \leftarrow x_i^{2}$
   \ENDIF
   \ENDFOR
   \RETURN $s^o=(x_1^{o},x_2^{o},\dots,x_n^{o})$
 \end{algorithmic}
 \end{small}
\end{algorithm}

\begin{algorithm}[!htbp]
\begin{small}
 \caption{The Greedy Crossover Operator for MaxMeanDP} \label{Algo_greedy_crossover}
 \begin{algorithmic}[1]
   \STATE \sf \textbf{Input}: Two parent solutions $s^1$ and $s^2$, and their subsets of selected elements are respectively denoted by $M_1$ and $M_2$
   \STATE \textbf{Output}:  Offspring solution $s^o$ whose subset of selected elements is denoted by $M_o$  \\
   \STATE $m \leftarrow (|M_1|+|M_2|)/2$ \ \ \ /* Determine approximately the size of the set $M_o$ */ \\
    /*Generate a partial solution by preserving the common elements of $M_1$ and $M_2$ */
   \STATE $M_o \leftarrow M_1 \cap M_2 $
   \STATE $M_1^{'} \leftarrow M_1 \setminus M_o$,  $M_2^{'} \leftarrow M_2 \setminus M_o$
   \WHILE{$ |M_o| < m $ }
         \IF {$M_1^{'} \neq \emptyset$}
         \STATE $v_o \leftarrow argmax\{\Delta_f(v):v \in M_1^{'}\}$ \\  /* $\Delta_f(v)$ represents the move value of adding element $v$ to $M_o$ */
         \STATE $M_o \leftarrow M_o \cup \{v_o\}$, $M_1^{'} \leftarrow M_1^{'} \setminus \{v_o\}$
		 \ENDIF
         \IF {$M_2^{'} \neq \emptyset$}
         \STATE $u_o \leftarrow argmax\{\Delta_f(u):u \in M_2^{'}\}$
         \STATE $M_o \leftarrow M_o \cup \{u_o\}$, $M_2^{'} \leftarrow M_2^{'} \setminus \{u_o\}$
		 \ENDIF
   \ENDWHILE
   \RETURN $s^o$(i.e.,$M_o$)
 \end{algorithmic}
 \end{small}
\end{algorithm}

In a memetic algorithm, the crossover operator is another essential ingredient whose main goal is to bring the search process to new promising search regions to diversify the search. In this work, we investigate two crossover operators for MaxMeanDP; the first one is the standard uniform crossover (denoted by UC) operator, and the other is a specific greedy crossover (denoted by GC) operator.

UC is very simple and is described in Algorithm \ref{Algo_random_crossover}. Given two parent solutions $s^1=(x_1^{1},x_2^{1},\dots,x_n^{1})$ and $s^2=(x_1^{2},x_2^{2},\dots,x_n^{2})$, the value of each component $x_i^{o}$ ($i=1,2,\dots,n$) of the offspring solution $s^o$ is randomly chosen from the set $\{x_i^{1},x_i^{2}\}$ with the same probability of 0.5. In spite of its simplicity, UC has shown to be quite robust and effective in many settings.

The dedicated GC operator is shown in Algorithm \ref{Algo_greedy_crossover}. Given two parent solutions $s^1$ and $s^2$ whose subsets of selected elements are respectively represented by $M_1$ and $M_2$, we first estimate the size of the set $M_o$ of selected elements for the offspring solution $s^o$ such that the Hamming distances of $s^o$ to $s^1$ as well as $s^2$ are approximately the same, i.e., $m \leftarrow \frac{|M_1|+|M_2|}{2}$. Then a partial solution is generated by preserving the common elements of $M_1$ and $M_2$, i.e., $M_o \leftarrow M_1 \cap M_2$. After that, we complete the offspring solution $s^o$ in a step by step way by choosing in turn a best element from $M_1\setminus M_o$ or $M_2\setminus M_o$ and adding it to $M_o$. More specifically, we first choose an element ($v_o$) that yields the largest move value from $M_1 \setminus M_o$ and add it to $M_o$, then we choose another element ($u_o$) that yields the largest move value from $M_2 \setminus M_o$ and add it to $M_o$. We repeat these two steps until the size of $M_o$ reaches $m$.

Intuitively, UC is more disruptive than GC and thus is suitable for the purpose of diversifying the search. On the other hand, GC is more conservative and computationally more expensive. Based on the computational outcomes presented in Section \ref{comparison_co}, we adopt in this study the UC operator as the main crossover operator of our memetic algorithm, while using GC as a reference operator for our analysis on crossovers.

\subsection{Population Updating Rule}
\label{subsec_pool_updating}

When a new offspring solution is generated by the crossover operator, it is first improved the tabu search procedure and then used to update the population according to the following rule. If the offspring solution is distinct from any existing solution in the population and is better than the worst solution in the population in terms of objective value, then the offspring solution replaces the worst solution of the population. Otherwise, the population is kept unchanged.

\section{Experimental Results and Comparisons}
\label{results}

In this section, we run extensive computational experiments to assess the performance of our memetic algorithm based on a large number of MaxMeanDP benchmark instances.

\subsection{Benchmark Instances}
\label{instance}

Our computational experiments are carried out on two types of instances, namely Type I and Type II. The distances of Type I instances are randomly generated in the interval $[-10,10]$ with a uniform probability distribution, while the distances of Type II instances are generated from $[-10,-5] \cup [5,10]$ with the same probability distribution.

Additionally, the set of benchmark instances used in our experiments is composed of two subsets. The first subset consists of 80 Type I instances and 80 Type II instances with the number of elements $n$ ranging from 20 to 1000. These 160 instances were extensively adopted by the previous studies \cite{Carrasco2014,DellaCroce2014,Marti2013} and are available online at \url{http://www.optsicom.es}. The second subset consists of 20 Type I and 20 Type II large instances with $n=3000$ or 5000. The source code of the generator used to obtain these 40 large instances will be available\footnote{The source code of generating these instances will be available from our website.}.

\subsection{Parameter Settings and Experimental Protocol}
\label{exprimental_Protocol}

Our memetic algorithm relies on only three parameters: the population size $p$, the depth of tabu search $\alpha$ and the maximum tabu tenure $T_{max}$. For $p$ and $\alpha$, we follow \cite{Wu2013b} and set $p=10, \alpha=50000$ while setting $T_{max} = 120$ empirically. This parameter setting is used for all the experiments reported in the paper. Even if fine-tuning these parameters would lead to better results, as we show below, our algorithm with this fixed setting is able to attain a high performance with respect to the state of the art results.

Our memetic algorithm is programmed in C++ and compiled using g++ compiler with the '-O2' flag\footnote{Our best results and the source code of our algorithm will be made available online.}. All experiments are carried out on a computer with an Intel Xeon E5440 processor (2.83 GHz CPU and 2Gb RAM), running the Linux operating system.  Following the DIMACS machine benchmark procedure\footnote{dmclique, ftp://dimacs.rutgers.edu/pub/dsj/clique, the benchmark procedure is complied by gcc compiler with the '-O2' flag}, our machine requires respectively 0.23, 1.42, and 5.42 seconds for the graphs r300.5, r400.5, r500.5.

Given the stochastic nature of our algorithm, we solve each tested problem instance 20 times, where the stopping condition is given by a cutoff time limit which depends on the size of the instances. Specifically, the cutoff limit $t_{out}$ is set to be 10 seconds for $n\le 150$, 100 seconds for $n \in [500,1000]$, 1000 seconds for $n=3000$, and 2000 seconds for $n=5000$. As we discuss in Section \ref{resutls_small_instances}, these time limits are significantly shorter than those used by the reference algorithms of the literature.

\subsection{Computational Results and Comparisons on Small and Medium Sized Instances}
\label{resutls_small_instances}

\renewcommand{\baselinestretch}{0.6}\huge\normalsize
\begin{table}[!htp]\centering
\caption{Computational results of the proposed MAMMDP algorithm on the set of 60 representative instances with $500 \le n\le1000$. Each instance is independently solved 20 times, and improved results are indicated in bold compared to the previous best known results $f_{pre}$ of the literature reported in \cite{Carrasco2014,DellaCroce2014,Marti2013}.} \label{Results_MA_Small}
\begin{tiny}
\begin{tabular}{p{1.8cm}p{0.8cm}p{0.8cm}p{0.01cm}p{1.5cm}p{1.5cm}p{0.8cm}p{0.8cm}p{0.01cm}}
\hline
  &   &   &   & \multicolumn{4}{c}{Memetic Algorithm}  & \\
\cline{5-8}
Instance  &  n  & \centering{$f_{pre}$ \cite{Carrasco2014,DellaCroce2014,Marti2013}} &   & $f_{best}$ & $f_{avg}$ & SR & $t(s)$ & \\
\hline
MDPI1\_500    & 500  & 81.28  &  & 81.277044  & 81.277044  & 20/20 & 0.69  \\
MDPI2\_500    & 500  & 77.60  &  & \textbf{78.610216}  & \textbf{78.610216}  & 20/20 & 1.43  \\
MDPI3\_500    & 500  & 75.65  &  & \textbf{76.300787}  & \textbf{76.300787}  & 20/20 & 2.71  \\
MDPI4\_500    & 500  & 82.28  &  & \textbf{82.332081}  & \textbf{82.332081}  & 20/20 & 0.95  \\
MDPI5\_500    & 500  & 80.01  &  & \textbf{80.354029}  & \textbf{80.354029}  & 20/20 & 2.80  \\
MDPI6\_500    & 500  & 81.12  &  & \textbf{81.248553}  & \textbf{81.248553}  & 20/20 & 0.78  \\
MDPI7\_500    & 500  & 78.09  &  & \textbf{78.164511}  & \textbf{78.164511}  & 20/20 & 0.92  \\
MDPI8\_500    & 500  & 79.01  &  & \textbf{79.139881}  & \textbf{79.139881}  & 20/20 & 1.27  \\
MDPI9\_500    & 500  & 77.15  &  & \textbf{77.421000}  & \textbf{77.421000}  & 20/20 & 2.37  \\
MDPI10\_500   & 500  & 81.24  &  & \textbf{81.309871}  & \textbf{81.309871}  & 20/20 & 0.91  \\
MDPII1\_500   & 500  & 109.33 &  & \textbf{109.610136} & \textbf{109.610136} & 20/20 & 0.75  \\
MDPII2\_500   & 500  & 105.06 &  & \textbf{105.717536} & \textbf{105.717536} & 20/20 & 0.88  \\
MDPII3\_500   & 500  & 107.64 &  & \textbf{107.821739} & \textbf{107.821739} & 20/20 & 0.89  \\
MDPII4\_500   & 500  & 105.69 &  & \textbf{106.100071} & \textbf{106.100071} & 20/20 & 0.56  \\
MDPII5\_500   & 500  & 106.59 &  & \textbf{106.857162} & \textbf{106.857162} & 20/20 & 0.99  \\
MDPII6\_500   & 500  & 106.17 &  & \textbf{106.297958} & \textbf{106.297958} & 20/20 & 0.98  \\
MDPII7\_500   & 500  & 106.92 &  & \textbf{107.149379} & \textbf{107.149379} & 20/20 & 0.88  \\
MDPII8\_500   & 500  & 103.49 &  & \textbf{103.779195} & \textbf{103.779195} & 20/20 & 0.59  \\
MDPII9\_500   & 500  & 106.20 &  & \textbf{106.619793} & \textbf{106.619793} & 20/20 & 1.11  \\
MDPII10\_500  & 500  & 103.79 &  & \textbf{104.651507} & \textbf{104.651507} & 20/20 & 1.01  \\
MDPI1\_750    & 750  & 95.86  &  & \textbf{96.644236}  & \textbf{96.644236}  & 20/20 & 7.98  \\
MDPI2\_750    & 750  & 97.42  &  & \textbf{97.564880}  & \textbf{97.564880}   & 20/20 & 3.82  \\
MDPI3\_750    & 750  & 96.97  &  & \textbf{97.798864}  & \textbf{97.798864}  & 20/20 & 1.81  \\
MDPI4\_750    & 750  & 95.21  &  & \textbf{96.041364}  & \textbf{96.041364}  & 20/20 & 4.38  \\
MDPI5\_750    & 750  & 96.65  &  & \textbf{96.740448}  & \textbf{96.740448}  & 20/20 & 1.21  \\
MDPI6\_750    & 750  & 99.25  &  & \textbf{99.861250}  & \textbf{99.861250}  & 20/20 & 5.55  \\
MDPI7\_750    & 750  & 96.26  &  & \textbf{96.545413}  & \textbf{96.545413}  & 20/20 & 1.01  \\
MDPI8\_750    & 750  & 96.46  &  & \textbf{96.726976}  & \textbf{96.726976}  & 20/20 & 1.73  \\
MDPI9\_750    & 750  & 96.78  &  & \textbf{98.058377}  & \textbf{98.058377}  & 20/20 & 2.18  \\
MDPI10\_750   & 750  & 99.85  &  & \textbf{100.064185} & \textbf{100.064185} & 20/20 & 3.42  \\
MDPII1\_750   & 750  & 127.69 &  & \textbf{128.863707} & \textbf{128.863707} & 20/20 & 5.66  \\
MDPII2\_750   & 750  & 130.79 &  & \textbf{130.954426} & \textbf{130.954426} & 20/20 & 2.31  \\
MDPII3\_750   & 750  & 129.40 &  & \textbf{129.782453} & \textbf{129.782453} & 20/20 & 11.64 \\
MDPII4\_750   & 750  & 125.68 &  & \textbf{126.582271} & \textbf{126.582271} & 20/20 & 1.48  \\
MDPII5\_750   & 750  & 128.13 &  & \textbf{129.122878} & \textbf{129.122878} & 20/20 & 1.32  \\
MDPII6\_750   & 750  & 128.55 &  & \textbf{129.025215} & \textbf{129.025215} & 20/20 & 7.98  \\
MDPII7\_750   & 750  & 124.91 &  & \textbf{125.646682} & \textbf{125.646682} & 20/20 & 3.38  \\
MDPII8\_750   & 750  & 130.66 &  & \textbf{130.940548} & \textbf{130.940548} & 20/20 & 1.91  \\
MDPII9\_750   & 750  & 128.89 &  & \textbf{128.889908} & \textbf{128.889908} & 20/20 & 1.30  \\
MDPII10\_750  & 750  & 132.99 &  & \textbf{133.265300} & \textbf{133.265300} & 20/20 & 1.81  \\
MDPI1\_1000   & 1000 & 118.76 &  & \textbf{119.174112} & \textbf{119.174112} & 20/20 & 8.25  \\
MDPI2\_1000   & 1000 & 113.22 &  & \textbf{113.524795} & \textbf{113.524795} & 20/20 & 3.52  \\
MDPI3\_1000   & 1000 & 114.51 &  & \textbf{115.138638} & \textbf{115.138638} & 20/20 & 2.32  \\
MDPI4\_1000   & 1000 & 110.53 &  & \textbf{111.127437} & \textbf{111.127437} & 20/20 & 5.72  \\
MDPI5\_1000   & 1000 & 111.24 &  & \textbf{112.723188} & \textbf{112.723188} & 20/20 & 1.61  \\
MDPI6\_1000   & 1000 & 112.08 &  & \textbf{113.198718} & \textbf{113.198718} & 20/20 & 7.72  \\
MDPI7\_1000   & 1000 & 110.94 &  & \textbf{111.555536} & \textbf{111.555536} & 20/20 & 1.88  \\
MDPI8\_1000   & 1000 & 110.29 &  & \textbf{111.263194} & \textbf{111.263194} & 20/20 & 3.55  \\
MDPI9\_1000   & 1000 & 115.78 &  & \textbf{115.958833} & \textbf{115.958833} & 20/20 & 2.38  \\
MDPI10\_1000  & 1000 & 114.29 &  & \textbf{114.731644} & \textbf{114.731644} & 20/20 & 2.16  \\
MDPII1\_1000  & 1000 & 145.46 &  & \textbf{147.936175} & \textbf{147.936175} & 20/20 & 1.60  \\
MDPII2\_1000  & 1000 & 150.49 &  & \textbf{151.380035} & \textbf{151.380035} & 20/20 & 1.78  \\
MDPII3\_1000  & 1000 & 149.36 &  & \textbf{150.788178} & \textbf{150.788178} & 20/20 & 4.92  \\
MDPII4\_1000  & 1000 & 147.91 &  & \textbf{149.178006} & \textbf{149.178006} & 20/20 & 3.80  \\
MDPII5\_1000  & 1000 & 150.23 &  & \textbf{151.520847} & \textbf{151.520847} & 20/20 & 3.28  \\
MDPII6\_1000  & 1000 & 147.29 &  & \textbf{148.343378} & \textbf{148.343378} & 20/20 & 3.22  \\
MDPII7\_1000  & 1000 & 148.41 &  & \textbf{148.742375} & \textbf{148.742375} & 20/20 & 6.30  \\
MDPII8\_1000  & 1000 & 145.87 &  & \textbf{147.826804} & \textbf{147.826804} & 20/20 & 13.52 \\
MDPII9\_1000  & 1000 & 145.67 &  & \textbf{147.078145} & \textbf{147.078145} & 20/20 & 1.63  \\
MDPII10\_1000 & 1000 & 148.40 &  & \textbf{150.046137} & \textbf{150.046137} & 20/20 & 2.13 \\
\hline
\#Better      &     &        &  & \centering{59}& \centering{59} &  &  &  \\
\#Equal       &     &        &  & \centering{1} & \centering{1}  &  &  &  \\
\#Worse       &     &        &  & \centering{0} & \centering{0}  &  &  &  \\
\hline
\end{tabular}
\end{tiny}
\end{table}
\renewcommand{\baselinestretch}{1.0}\large\normalsize

Our first experiment aims to evaluate the performance of our MAMMDP algorithm on the set of 160 popular instances with up to $1000$ elements. The computational results of MAMMDP on the 60 medium sized instances are summarized in Table \ref{Results_MA_Small}, whereas the results of the 100 small instances with $n\le 150$ are given in the Appendix \ref{Results_MA_Very_Small} with the same computational statistics.

The first two columns of the table give respectively the name and size of instances. Column 3 indicates the best objective values ($f_{pre}$) of the literature which are compiled from the best results yielded by three recent and best performing algorithms, namely GRASP-PR \cite{Marti2013}, a hybrid heuristic approach \cite{DellaCroce2014}, and a diversified tabu search method \cite{Carrasco2014} (from \url{http://www.optsicom.es}). Note that the previous best known results ($f_{pre}$) are given with two decimal in the literature. In \cite{Carrasco2014,Marti2013}, the cutoff time limits are set to 90, 600, and 1800 seconds for instances with the size 500, 750, and 1000, respectively, and in \cite{DellaCroce2014} the cutoff time limits are respectively set to 60 and 600 seconds for the instances with the size 150 and 500. The GRASP-PR method was performed on a computer with an Intel Core Solo 1.4 GHz CPU with 3 GB RAM \cite{Marti2013}; the hybrid heuristic approach was run on a computer with an Intel Core i5-3550 3.30GHz CPU with 4GB RAM \cite{DellaCroce2014} and the diversified tabu search method was run on a computer with an Intel Core 2 Quad CPU and 6 GB RAM \cite{Carrasco2014}.

Our results are reported in columns 4 to 7, including the best objective value ($f_{best}$) yielded over 20 independent runs, the average objective value ($f_{avg}$), the success rate ($SR$) to achieve $f_{best}$, and the average computing time in seconds ($t(s)$) to achieve $f_{best}$. The last three rows \emph{Better, Equal, Worse} of the table respectively show the number of instances for which our result is better, equal to and worse than $f_{pre}$. The improved results are indicated in bold compared to $f_{pre}$.

First, one observes from Table \ref{Results_MA_Small} that our MAMMDP algorithm improves the previous best known result for all instances except for one instance for which our result matches the previous best known result. Therefore, these results clearly indicate the superiority of the proposed MAMMDP algorithm compared to the previous MaxMeanDP algorithms. Second, when examining the success rate of the algorithm, one can find that the MAMMDP algorithm achieves a success rate of 100\% for all tested instances, which means a good robustness of our MAMMDP algorithm. Third, in terms of average computing time, one observes that for all instances our MAMMDP algorithm obtains its best result with an average time of less than 14 seconds, which are much shorter than those of the previous algorithms in the literature.

\subsection{Computational Results on Large-Scale Instances}
\label{result_large_instance}

\renewcommand{\baselinestretch}{0.7}\huge\normalsize
\begin{table}[!htp]\centering
\caption{Computational results of the proposed memetic algorithm on the set of 40 large instances with $n=3000, 5000$. Each instance is independently solved 20 times.} \label{Results_MA_Large}
\begin{tiny}
\begin{tabular}{p{1.8cm}p{0.8cm}p{0.01cm}p{1.5cm}p{1.5cm}p{0.8cm}p{0.8cm}p{0.01cm}}
\hline
  &   &    & \multicolumn{4}{c}{Memetic Algorithm}  & \\
\cline{4-7}
Instance  &  n  &   & $f_{best}$ & $f_{avg}$ & SR & $t(s)$ & \\
\hline
MDPI1\_3000   & 3000 &  & 189.048965 & 189.048965 & 20/20 & 88.36   \\
MDPI2\_3000   & 3000 &  & 187.387292 & 187.387292 & 20/20 & 60.71   \\
MDPI3\_3000   & 3000 &  & 185.666806 & 185.655084 & 13/20 & 352.85  \\
MDPI4\_3000   & 3000 &  & 186.163727 & 186.153631 & 16/20 & 300.37  \\
MDPI5\_3000   & 3000 &  & 187.545515 & 187.545515 & 20/20 & 61.29   \\
MDPI6\_3000   & 3000 &  & 189.431257 & 189.431257 & 20/20 & 51.99   \\
MDPI7\_3000   & 3000 &  & 188.242583 & 188.242583 & 20/20 & 86.57   \\
MDPI8\_3000   & 3000 &  & 186.796814 & 186.796814 & 20/20 & 48.04   \\
MDPI9\_3000   & 3000 &  & 188.231264 & 188.231264 & 20/20 & 151.78  \\
MDPI10\_3000  & 3000 &  & 185.682511 & 185.623778 & 10/20 & 228.72  \\
MDPII1\_3000  & 3000 &  & 252.320433 & 252.320433 & 20/20 & 59.70   \\
MDPII2\_3000  & 3000 &  & 250.062137 & 250.062137 & 20/20 & 220.10  \\
MDPII3\_3000  & 3000 &  & 251.906270 & 251.906270 & 20/20 & 146.32  \\
MDPII4\_3000  & 3000 &  & 253.941007 & 253.940596 & 19/20 & 370.76  \\
MDPII5\_3000  & 3000 &  & 253.260423 & 253.260350 & 17/20 & 374.00  \\
MDPII6\_3000  & 3000 &  & 250.677750 & 250.677750 & 20/20 & 55.35   \\
MDPII7\_3000  & 3000 &  & 251.134413 & 251.134413 & 20/20 & 74.72   \\
MDPII8\_3000  & 3000 &  & 252.999648 & 252.999648 & 20/20 & 79.82   \\
MDPII9\_3000  & 3000 &  & 252.425770 & 252.425770 & 20/20 & 90.27   \\
MDPII10\_3000 & 3000 &  & 252.396590 & 252.396590 & 20/20 & 13.18   \\
MDPI1\_5000   & 5000 &  & 240.162535 & 240.102875 & 7/20  & 312.13  \\
MDPI2\_5000   & 5000 &  & 241.827401 & 241.792978 & 6/20  & 1244.36 \\
MDPI3\_5000   & 5000 &  & 240.890819 & 240.888162 & 19/20 & 810.48  \\
MDPI4\_5000   & 5000 &  & 240.997186 & 240.976789 & 6/20  & 653.64  \\
MDPI5\_5000   & 5000 &  & 242.480129 & 242.475885 & 19/20 & 735.16  \\
MDPI6\_5000   & 5000 &  & 240.322850 & 240.306326 & 8/20  & 976.02  \\
MDPI7\_5000   & 5000 &  & 242.814943 & 242.774982 & 5/20  & 259.50  \\
MDPI8\_5000   & 5000 &  & 241.194990 & 241.161763 & 8/20  & 1148.60 \\
MDPI9\_5000   & 5000 &  & 239.760560 & 239.667613 & 4/20  & 1219.71 \\
MDPI10\_5000  & 5000 &  & 243.473734 & 243.373015 & 4/20  & 457.28  \\
MDPII1\_5000  & 5000 &  & 322.235897 & 322.181291 & 5/20  & 1519.05 \\
MDPII2\_5000  & 5000 &  & 327.301910 & 327.006342 & 5/20  & 1103.13 \\
MDPII3\_5000  & 5000 &  & 324.813456 & 324.801590 & 10/20 & 955.81  \\
MDPII4\_5000  & 5000 &  & 322.237586 & 322.197276 & 5/20  & 664.10  \\
MDPII5\_5000  & 5000 &  & 322.491211 & 322.380726 & 7/20  & 1014.90 \\
MDPII6\_5000  & 5000 &  & 322.950488 & 322.703887 & 4/20  & 352.88  \\
MDPII7\_5000  & 5000 &  & 322.850438 & 322.793125 & 10/20 & 714.31  \\
MDPII8\_5000  & 5000 &  & 323.112120 & 323.053268 & 11/20 & 879.48  \\
MDPII9\_5000  & 5000 &  & 323.543775 & 323.339842 & 7/20  & 569.73  \\
MDPII10\_5000 & 5000 &  & 324.519908 & 324.414458 & 15/20 & 752.95  \\
\hline
\end{tabular}
\end{tiny}
\end{table}
\renewcommand{\baselinestretch}{1.0}\large\normalsize

In order to further assess the performance of the proposed MAMMDP algorithm on the large-scale instances, we run independently  MAMMDP 20 times to solve each instance of the second set of benchmarks with $n\ge3000$, and report the computational statistics in Table \ref{Results_MA_Large} with the same information as in Table \ref{Results_MA_Small}.

Table \ref{Results_MA_Large} discloses that for the instances with 3000 elements, our MAMMDP algorithm reaches a success rate of at least 10/20, which is an interesting indicator as to its good performance for these instances. However, for the still larger instances with $n=5000$, the success rate of the algorithm significantly varies between 4/20 and 19/20, which means that these large instances are clearly more difficult. Nevertheless, one observes that the difference between the best and average objective values is very small for all instances. These results can be served as reference lower bounds for future comparisons of new MaxMeanDP algorithms.

\section{Analysis and Discussions}
\label{Discussion}

In this section, we study some essential ingredients of the proposed algorithm to understand their impacts on the performance of the proposed algorithm.

\subsection{Effectiveness of the Tabu Search Procedure }
\label{Discussion_TS}

\renewcommand{\baselinestretch}{0.8}\huge\normalsize
\begin{table}[!htp]\centering
\caption{Computational results of the underlying tabu search procedure of the proposed algorithm on the set of 30 representative instances with $500 \le n\le1000$. Each instance is independently solved 20 times, and improved results are indicated in bold compared to the previous best known objective values.} \label{Results_TS}
\begin{tiny}
\begin{tabular}{p{1.8cm}p{0.5cm}p{0.8cm}p{1.0cm}p{0.01cm}p{1.5cm}p{1.5cm}p{0.6cm}p{0.6cm}p{0.01cm}}
\hline
  &   &   &   &   & \multicolumn{4}{c}{Tabu Search}  & \\
\cline{6-9}
Instance  &  n  & \centering{$f_{pre}$} &  \centering{$f^{*}$}  &   & $f_{best}$ & $f_{avg}$ & SR & $t(s)$ & \\
\hline
MDPI1\_500   & 500  & 81.28  & 81.277044  &  & 81.277044  & 81.277044  & 20/20 & 0.54 \\
MDPI2\_500   & 500  & 77.60  & 78.610216  &  & \textbf{78.610216}  & \textbf{78.586017}  & 8/20  & 0.68 \\
MDPI3\_500   & 500  & 75.65  & 76.300787  &  & \textbf{76.300787}  & \textbf{76.252735}  & 7/20  & 0.80 \\
MDPI4\_500   & 500  & 82.28  & 82.332081  &  & \textbf{82.332081}  & \textbf{82.325976}  & 15/20 & 0.65 \\
MDPI5\_500   & 500  & 80.01  & 80.354029  &  & \textbf{80.354029}  & \textbf{80.349285}  & 5/20  & 0.78 \\
MDPII1\_500  & 500  & 109.33 & 109.610136 &  & \textbf{109.610136} & \textbf{109.473283} & 15/20 & 0.56 \\
MDPII2\_500  & 500  & 105.06 & 105.717536 &  & \textbf{105.717536} & \textbf{105.678373} & 14/20 & 0.70 \\
MDPII3\_500  & 500  & 107.64 & 107.821739 &  & \textbf{107.821739} & \textbf{107.808748} & 18/20 & 0.77 \\
MDPII4\_500  & 500  & 105.69 & 106.100071 &  & \textbf{106.100071} & \textbf{106.08738}  & 18/20 & 0.54 \\
MDPII5\_500  & 500  & 106.59 & 106.857162 &  & \textbf{106.857162} & \textbf{106.834002} & 14/20 & 0.64 \\
MDPI1\_750   & 750  & 95.86  & 96.644236  &  & \textbf{96.644236}  & \textbf{96.472938}  & 5/20  & 1.24 \\
MDPI2\_750   & 750  & 97.42  & 97.564880  &  & \textbf{97.564880}   & \textbf{97.557159}  & 9/20  & 1.10 \\
MDPI3\_750   & 750  & 96.97  & 97.798864  &  & \textbf{97.798864}  & \textbf{97.780265}  & 15/20 & 1.05 \\
MDPI4\_750   & 750  & 95.21  & 96.041364  &  & \textbf{96.041364}  & \textbf{95.880611}  & 2/20  & 1.17 \\
MDPI5\_750   & 750  & 96.65  & 96.740448  &  & \textbf{96.740448}  & \textbf{96.731085}  & 18/20 & 0.99 \\
MDPII1\_750  & 750  & 127.69 & 128.863707 &  & \textbf{128.863707} & \textbf{128.401188} & 4/20  & 1.12 \\
MDPII2\_750  & 750  & 130.79 & 130.954426 &  & \textbf{130.954426} & \textbf{130.920628} & 7/20  & 1.01 \\
MDPII3\_750  & 750  & 129.40  & 129.782453 &  & \textbf{129.782453} & \textbf{129.557185} & 1/20  & 1.03 \\
MDPII4\_750  & 750  & 125.68 & 126.582271 &  & \textbf{126.582271} & \textbf{126.441529} & 11/20 & 0.96 \\
MDPII5\_750  & 750  & 128.13 & 129.122878 &  & \textbf{129.122878} & \textbf{129.032812} & 15/20 & 1.11 \\
MDPI1\_1000  & 1000 & 118.76 & 119.174112 &  & \textbf{119.174112} & \textbf{118.836369} & 5/20  & 1.63 \\
MDPI2\_1000  & 1000 & 113.22 & 113.524795 &  & \textbf{113.524795} & \textbf{113.363683} & 7/20  & 1.74 \\
MDPI3\_1000  & 1000 & 114.51 & 115.138638 &  & \textbf{115.138638} & \textbf{115.106057} & 16/20 & 1.33 \\
MDPI4\_1000  & 1000 & 110.53 & 111.127437 &  & \textbf{111.127437} & \textbf{110.924380} & 4/20  & 1.42 \\
MDPI5\_1000  & 1000 & 111.24 & 112.723188 &  & \textbf{112.723188} & \textbf{112.723188} & 20/20 & 1.18 \\
MDPII1\_1000 & 1000 & 145.46 & 147.936175 &  & \textbf{147.936175} & \textbf{147.919913} & 17/20 & 1.44 \\
MDPII2\_1000 & 1000 & 150.49 & 151.380035 &  & \textbf{151.380035} & \textbf{151.344939} & 18/20 & 1.54 \\
MDPII3\_1000 & 1000 & 149.36 & 150.788178 &  & \textbf{150.788178} & \textbf{150.475002} & 7/20  & 1.82 \\
MDPII4\_1000 & 1000 & 147.91 & 149.178006 &  & \textbf{149.178006} & \textbf{149.093037} & 11/20 & 1.59 \\
MDPII5\_1000 & 1000 & 150.23 & 151.520847 &  & \textbf{151.520847} & \textbf{151.495094} & 10/20 & 1.57 \\
\hline
\#Better      &     &    &    &   & \centering{29}& \centering{29} &  &  &  \\
\#Equal       &     &    &    &   & \centering{1} & \centering{1}  &  &  &  \\
\#Worse       &     &    &    &   & \centering{0} & \centering{0}  &  &  &  \\
\hline
\end{tabular}
\end{tiny}
\end{table}
\renewcommand{\baselinestretch}{1.0}\large\normalsize

First, to show the effectiveness of the underlying tabu search procedure of the proposed algorithm, we carry out an additional experiment on 30 representative instances, where each instance is independently solved 20 times by the tabu search procedure with randomly generated initial solutions. The computational results are summarized in Table \ref{Results_TS}, where $f^{*}$ and $t(s)$ represent respectively the previous best known result and the average computing time over 20 runs, other symbols are the same as those in Table \ref{Results_MA_Small}. Note that the improved results compared to the previous best known values $f_{pre}$ are indicated in bold.

Table \ref{Results_TS} discloses that our tabu search procedure alone is able to attain the previous best known result for the tested instances. Moreover, for all tested instances, the obtained average objective value ($f_{avg}$) is still better than the previous best known result ($f_{pre}$) on 29 out of 30 instances. These results indicate that our tabu search method is quite effective with respect to the existing algorithms designed for MaxMeanDP. In terms of computational time, one observes that one run of the tabu search procedure takes on average less than 2 seconds for instances with $n\le1000$, much shorter than those required by the state of the art algorithms \cite{Carrasco2014,DellaCroce2014,Marti2013}. To sum, this experiment shows that our tabu search procedure is highly effective compared with the existing state-of-the-art algorithms in the literature.

It should be mentioned that compared to the previous methods for MaxMeanDP, such as those in \cite{Carrasco2014}, the success of our tabu search method may be attributed to the combined use of the neighborhood $N_1$, the fast neighborhood evaluation technique and its dynamic tabu list management strategy.

\subsection{Influence of the Crossover Operator}
\label{comparison_co}

\renewcommand{\baselinestretch}{0.7}\huge\normalsize
\begin{table}[!htp]\centering
\caption{Comparison between the UC and GC crossover operators on the set of 40 large instances with $n=3000$ or $5000$. Each instance is independently solved 20 times by the memetic algorithms with the UC and GC operators respectively, and better results between these two memetic algorithms are indicated in bold .} \label{Results_comparison_co}
\begin{tiny}
\begin{tabular}{p{1.2cm}p{0.01cm}p{1.4cm}p{1.4cm}p{0.5cm}p{0.6cm}p{0.01cm}p{1.4cm}p{1.4cm}p{0.5cm}p{0.6cm}p{0.01cm}}
\hline
               &  &  \multicolumn{4}{c}{Uniform Crossover}     &  &    \multicolumn{4}{c}{Greedy Crossover}  & \\
\cline{3-6} \cline{8-11}
Instance       &  & $f_{best}$ & $f_{avg}$  & SR    & $t(s)$  &  &$f_{best}$  & $f_{avg}$  & SR    & $t(s)$ & \\
\hline
MDPI1\_3000    &  & 189.048965 & 189.048965 & 20/20 & 88.36   &  & 189.048965 & 189.048965 & 20/20 & 82.49  & \\
MDPI2\_3000    &  & 187.387292 & 187.387292 & 20/20 & 60.71   &  & 187.387292 & 187.387292 & 20/20 & 71.95  & \\
MDPI3\_3000    &  & 185.666806 & 185.655084 & 13/20 & 352.85  &  & 185.666806 & \textbf{185.656214} & 14/20 & 430.55 & \\
MDPI4\_3000    &  & 186.163727 & 186.153631 & 16/20 & 300.37  &  & 186.163727 & \textbf{186.163727} & 20/20 & 248.47 & \\
MDPI5\_3000    &  & 187.545515 & 187.545515 & 20/20 & 61.29   &  & 187.545515 & 187.545515 & 20/20 & 87.26  & \\
MDPI6\_3000    &  & 189.431257 & 189.431257 & 20/20 & 51.99   &  & 189.431257 & 189.431257 & 20/20 & 51.15  & \\
MDPI7\_3000    &  & 188.242583 & 188.242583 & 20/20 & 86.57   &  & 188.242583 & 188.242583 & 20/20 & 103.65 & \\
MDPI8\_3000    &  & 186.796814 & 186.796814 & 20/20 & 48.04   &  & 186.796814 & 186.796814 & 20/20 & 68.57  & \\
MDPI9\_3000    &  & 188.231264 & 188.231264 & 20/20 & 151.78  &  & 188.231264 & 188.231264 & 20/20 & 69.86  & \\
MDPI10\_3000   &  & 185.682511 & \textbf{185.623778} & 10/20 & 228.72  &  & 185.682511 & 185.623041 & 14/20 & 386.07 & \\
MDPII1\_3000   &  & 252.320433 & 252.320433 & 20/20 & 59.70   &  & 252.320433 & 252.320433 & 20/20 & 78.92  & \\
MDPII2\_3000   &  & 250.062137 & 250.062137 & 20/20 & 220.10  &  & 250.062137 & 250.062137 & 20/20 & 219.11 & \\
MDPII3\_3000   &  & 251.906270 & 251.906270 & 20/20 & 146.32  &  & 251.906270 & 251.906270 & 20/20 & 139.13 & \\
MDPII4\_3000   &  & 253.941007 & \textbf{253.940596} & 19/20 & 370.76  &  & 253.941007 & 253.938635 & 19/20 & 243.89 & \\
MDPII5\_3000   &  & 253.260423 & 253.260350 & 17/20 & 374.00  &  & 253.260423 & \textbf{253.260423} & 20/20 & 301.14 & \\
MDPII6\_3000   &  & 250.677750 & 250.677750 & 20/20 & 55.35   &  & 250.677750 & 250.677750 & 20/20 & 63.01  & \\
MDPII7\_3000   &  & 251.134413 & 251.134413 & 20/20 & 74.72   &  & 251.134413 & 251.134413 & 20/20 & 75.89  & \\
MDPII8\_3000   &  & 252.999648 & 252.999648 & 20/20 & 79.82   &  & 252.999648 & 252.999648 & 20/20 & 62.52  & \\
MDPII9\_3000   &  & 252.425770 & 252.425770 & 20/20 & 90.27   &  & 252.425770 & 252.425770 & 20/20 & 96.89  & \\
MDPII10\_3000  &  & 252.396590 & 252.396590 & 20/20 & 13.18   &  & 252.396590 & 252.396590 & 20/20 & 18.34  & \\
MDPI1\_5000    &  & 240.162535 & 240.102875 & 7/20  & 312.13  &  & 240.162535 & \textbf{240.110086} & 9/20  & 104.32 & \\
MDPI2\_5000    &  & 241.827401 & \textbf{241.792978} & 6/20  & 1244.36 &  & 241.827401 & 241.750738 & 3/20  & 873.37 & \\
MDPI3\_5000    &  & 240.890819 & \textbf{240.888162} & 19/20 & 810.48  &  & 240.890819 & 240.875111 & 14/20 & 681.51 & \\
MDPI4\_5000    &  & 240.997186 & 240.976789 & 6/20  & 653.64  &  & 240.997186 & \textbf{240.987885} & 10/20 & 1062.58& \\
MDPI5\_5000    &  & 242.480129 & \textbf{242.475885} & 19/20 & 735.16  &  & 242.480129 & 242.475876 & 18/20 & 790.37 & \\
MDPI6\_5000    &  & 240.322850 & \textbf{240.306326} & 8/20  & 976.02  &  & \textbf{240.376038} & 240.296124 & 4/20  & 564.79 & \\
MDPI7\_5000    &  & 242.814943 & 242.774982 & 5/20  & 259.50  &  & \textbf{242.820139} & \textbf{242.782093} & 5/20  & 261.98 & \\
MDPI8\_5000    &  & 241.194990 & \textbf{241.161763} & 8/20  & 1148.60 &  & 241.194990 & 241.159647 & 10/20 & 1219.58& \\
MDPI9\_5000    &  & 239.760560 & \textbf{239.667613} & 4/20  & 1219.71 &  & 239.760560 & 239.569628 & 4/20  & 756.59 & \\
MDPI10\_5000   &  & \textbf{243.473734} & \textbf{243.373015} & 4/20  & 457.28  &  & 243.385487 & 243.356904 & 9/20  & 1057.73& \\
MDPII1\_5000   &  & 322.235897 & \textbf{322.181291} & 5/20  & 1519.05 &  & 322.235897 & 322.172704 & 5/20  & 541.47 & \\
MDPII2\_5000   &  & 327.301910 & \textbf{327.006342} & 5/20  & 1103.13 &  & 327.301910 & 326.953531 & 7/20  & 662.58 & \\
MDPII3\_5000   &  & 324.813456 & \textbf{324.801590} & 10/20 & 955.81  &  & 324.813456 & 324.790747 & 4/20  & 629.53 & \\
MDPII4\_5000   &  & 322.237586 & 322.197276 & 5/20  & 664.10  &  & 322.237586 & \textbf{322.205163} & 7/20  & 535.54 & \\
MDPII5\_5000   &  & 322.491211 & 322.380726 & 7/20  & 1014.90 &  & 322.491211 & \textbf{322.403539} & 8/20  & 1095.86& \\
MDPII6\_5000   &  & 322.950488 & \textbf{322.703887} & 4/20  & 352.88  &  & 322.950488 & 322.692191 & 6/20  & 457.37 & \\
MDPII7\_5000   &  & 322.850438 & \textbf{322.793125} & 10/20 & 714.31  &  & 322.850438 & 322.774304 & 12/20 & 631.95 & \\
MDPII8\_5000   &  & 323.112120 & \textbf{323.053268} & 11/20 & 879.48  &  & 323.112120 & 323.007841 & 13/20 & 764.65 & \\
MDPII9\_5000   &  & 323.543775 & \textbf{323.339842} & 7/20  & 569.73  &  & 323.543775 & 323.273630 & 5/20  & 305.89 & \\
MDPII10\_5000  &  & 324.519908 & 324.414458 & 15/20 & 752.95  &  & 324.519908 & \textbf{324.500667} & 19/20 & 659.21 & \\
\hline
\#Better       &  & \centering{1}      & \centering{16}           &   &        &  & \centering{2}    & \centering{9}  &  &   & \\
\#Equal        &  & \centering{37}     & \centering{15}           &   &        &  & \centering{37}   & \centering{15} &  &   & \\
\#Worse        &  & \centering{2}      & \centering{9}            &   &        &  & \centering{1}    & \centering{16} &  &   & \\
\textit{p-value}  &  & \centering{5.637e-1} & \centering{1.615e-1}   &   &        &  &                  &                &  &   & \\
\hline
\end{tabular}
\end{tiny}
\end{table}
\renewcommand{\baselinestretch}{1.0}\large\normalsize

In this section, we show a study about the influence of the crossover operator on the performance of the proposed algorithm by comparing the uniform crossover (UC) operator and the greedy crossover (GC) operator. The experiment is carried out on a set of 40 large instance with $n\ge 3000$, where the memetic algorithms with UC and GC are respectively performed 20 times for each of the tested instances. The computational results of the both algorithms are summarized in Table \ref{Results_comparison_co}, including the best ($f_{best}$) and average ($f_{avg}$) objective values obtained over 20 runs, the success rate ($SR$) and average computing time in seconds ($t(s)$) to achieve the associated $f_{best}$, where better results between these two algorithms are indicated in bold. The rows, \emph{Better, Equal, Worse} denote respectively the number of instances for which an algorithm yields better, equal, worse results compared to the other algorithm. Finally, to verify whether there exists a significant difference between the two crossover operators in terms of the best and average objective values, the \textit{p-values} from the non-parametric Friedman test are reported in the last row.

One observes from Table \ref{Results_comparison_co} that the UC and GC operators are comparable in the overall performance of the algorithm in terms of best and average results, which is confirmed by the Friedman test ($p$-values $>0.05$). First, in terms of the best objective value, UC and GC yield respectively better result on 1 and 2 instances compared to another operator. Second, in terms of the average objective value, it can be found the UC operator produces a better and worse result respectively for 16 and 9 instances. As to the success rate and computing time, these two crossover operators also achieve similar performances. This experiment demonstrates that there is no dominance of one crossover over the other. Instead, they are complementary to solve different instances. One interesting future study would be to investigate the ways of using jointly these two operators with the search algorithm.

\subsection{Improvement of Memetic Algorithm Over the Tabu Search Procedure}
\label{MA_vs_LS}

As shown in Section \ref{Discussion_TS}, our tabu search procedure is very competitive compared to the existing algorithms in the literature. So it is interesting to know whether our MAMMDP algorithm has a significant improvement over its underlying local optimization (the tabu search) procedure. To compare the performances of the MAMMDP algorithm and its underlying tabu search procedure, the multi-start tabu search (MTS) and MAMMDP algorithms are respectively performed 20 times for each of the 40 representative instances with $n=3000$ or 5000 under the same cutoff time limits given in Section \ref{exprimental_Protocol}. Notice that for MTS, the tabu search procedure is run in a multi-start way with a randomly generated initial solution for each re-start until the timeout limit is reached, the tabu search procedure being re-started once the depth of tabu search $\alpha$ (which is set to $5\times 10^{4}$) is reached. The computational results of both algorithms are respectively summarized in Table \ref{results_LS_vs_MA} which is composed of two parts, where the symbols are the same with those in Table \ref{Results_comparison_co}.

\renewcommand{\baselinestretch}{0.7}\huge\normalsize
\begin{table}[!htp]\centering
\caption{Comparison between the multi-start tabu search method (MTS) and the proposed memetic algorithm on the set of 40 large instances with $n\ge3000$. Each instance is independently solved 20 times by both algorithms respectively, and better results between two algorithms are indicated in bold.} \label{results_LS_vs_MA}
\begin{tiny}
\begin{tabular}{p{1.2cm}p{0.01cm}p{1.4cm}p{1.4cm}p{0.5cm}p{0.6cm}p{0.01cm}p{1.4cm}p{1.4cm}p{0.5cm}p{0.6cm}p{0.01cm}}
\hline
               &  &  \multicolumn{4}{c}{Memetic Algorithm}     &  &    \multicolumn{4}{c}{MTS }  & \\
\cline{3-6} \cline{8-11}
Instance       &  & $f_{best}$ & $f_{avg}$  & SR    & $t(s)$  &  &$f_{best}$  & $f_{avg}$  & SR    & $t(s)$ & \\
\hline
MDPI1\_3000   &  & 189.048965 & 189.048965 & 20/20 & 88.36   &  & 189.048965 & 189.048965 & 20/20 & 120.05 \\
MDPI2\_3000   &  & 187.387292 & 187.387292 & 20/20 & 60.71   &  & 187.387292 & 187.387292 & 20/20 & 101.07 \\
MDPI3\_3000   &  & 185.666806 & \textbf{185.655084} & 13/20 & 352.85  &  & 185.666806 & 185.651588 & 10/20 & 526.05 \\
MDPI4\_3000   &  & 186.163727 & 186.153631 & 16/20 & 300.37  &  & 186.163727 & \textbf{186.163727} & 20/20 & 136.64 \\
MDPI5\_3000   &  & 187.545515 & 187.545515 & 20/20 & 61.29   &  & 187.545515 & 187.545515 & 20/20 & 133.70 \\
MDPI6\_3000   &  & 189.431257 & 189.431257 & 20/20 & 51.99   &  & 189.431257 & 189.431257 & 20/20 & 35.59  \\
MDPI7\_3000   &  & 188.242583 & 188.242583 & 20/20 & 86.57   &  & 188.242583 & 188.242583 & 20/20 & 137.56 \\
MDPI8\_3000   &  & 186.796814 & 186.796814 & 20/20 & 48.04   &  & 186.796814 & 186.796814 & 20/20 & 66.76  \\
MDPI9\_3000   &  & 188.231264 & 188.231264 & 20/20 & 151.78  &  & 188.231264 & 188.231264 & 20/20 & 101.11 \\
MDPI10\_3000  &  & 185.682511 & 185.623778 & 10/20 & 228.72  &  & 185.682511 & \textbf{185.672371} & 18/20 & 352.70 \\
MDPII1\_3000  &  & 252.320433 & 252.320433 & 20/20 & 59.70   &  & 252.320433 & 252.320433 & 20/20 & 72.49  \\
MDPII2\_3000  &  & 250.062137 & \textbf{250.062137} & 20/20 & 220.10  &  & 250.062137 & 250.054744 & 7/20  & 513.80 \\
MDPII3\_3000  &  & 251.906270  & 251.906270 & 20/20 & 146.32  &  & 251.906270  & 251.906270 & 20/20 & 127.32 \\
MDPII4\_3000  &  & 253.941007 & \textbf{253.940596} & 19/20 & 370.76  &  & 253.941007 & 253.939680 & 18/20 & 352.64 \\
MDPII5\_3000  &  & 253.260423 & \textbf{253.260350} & 17/20 & 374.00  &  & 253.260423 & 253.260164 & 14/20 & 349.19 \\
MDPII6\_3000  &  & 250.677750  & 250.677750 & 20/20 & 55.35   &  & 250.677750  & 250.677750 & 20/20 & 69.78  \\
MDPII7\_3000  &  & 251.134413 & 251.134413 & 20/20 & 74.72   &  & 251.134413 & 251.134413 & 20/20 & 97.74  \\
MDPII8\_3000  &  & 252.999648 & 252.999648 & 20/20 & 79.82   &  & 252.999648 & 252.999648 & 20/20 & 115.84 \\
MDPII9\_3000  &  & 252.425770  & 252.425770 & 20/20 & 90.27   &  & 252.425770  & 252.425770 & 20/20 & 106.79 \\
MDPII10\_3000 &  & 252.396590  & 252.396590 & 20/20 & 13.18   &  & 252.396590  & 252.396590 & 20/20 & 16.06  \\
\hline
\#Better       &  & \centering{0}     & \centering{4} &  \centering{4} &  \centering{14}     &  & \centering{0}    & \centering{2}   & \centering{2} & \centering{6}  & \\
\#Equal        &  & \centering{20}    & \centering{14}& \centering{14} &  \centering{0}      &  & \centering{20}   & \centering{14}  & \centering{14}& \centering{0}  & \\
\#Worse        &  & \centering{0}     & \centering{2} &  \centering{2} &  \centering{6}      &  & \centering{0}    & \centering{4}   & \centering{4} &  \centering{14} & \\
\textit{p-value}       &  & \centering{1.0}      & \centering{4.142e-1} &   &       &  &     &    &  &   & \\
\hline
MDPI1\_5000   &  & \textbf{240.162535} & \textbf{240.102875} & 7/20  & 312.13  &  & 240.141212 & 240.021201 & 2/20  & 163.67 \\
MDPI2\_5000   &  & \textbf{241.827401} & \textbf{241.792978} & 6/20  & 1244.36 &  & 241.817543 & 241.753546 & 2/20  & 734.33 \\
MDPI3\_5000   &  & 240.890819 & \textbf{240.888162} & 19/20 & 810.48  &  & 240.890819 & 240.825167 & 4/20  & 531.94 \\
MDPI4\_5000   &  & \textbf{240.997186} & \textbf{240.976789} & 6/20  & 653.64  &  & 240.973489 & 240.915459 & 3/20  & 560.43 \\
MDPI5\_5000   &  & 242.480129 & \textbf{242.475885} & 19/20 & 735.16  &  & 242.480129 & 242.430474 & 5/20  & 515.62 \\
MDPI6\_5000   &  & 240.322850  & \textbf{240.306326} & 8/20  & 976.02  &  & \textbf{240.328684} & 240.266264 & 5/20  & 290.72 \\
MDPI7\_5000   &  & 242.814943 & \textbf{242.774982} & 5/20  & 259.50  &  & \textbf{242.820139} & 242.759895 & 3/20  & 819.01 \\
MDPI8\_5000   &  & \textbf{241.194990}  & \textbf{241.161763} & 8/20  & 1148.60 &  & 241.144781 & 241.113453 & 3/20  & 721.59 \\
MDPI9\_5000   &  & 239.760560  & \textbf{239.667613} & 4/20  & 1219.71 &  & 239.760560  & 239.514958 & 4/20  & 360.05 \\
MDPI10\_5000  &  & \textbf{243.473734} & \textbf{243.373015} & 4/20  & 457.28  &  & 243.385487 & 243.348149 & 9/20  & 939.59 \\
MDPII1\_5000  &  & \textbf{322.235897} & \textbf{322.181291} & 5/20  & 1519.05 &  & 322.223220  & 322.131204 & 3/20  & 197.42 \\
MDPII2\_5000  &  & 327.301910  & 327.006342 & 5/20  & 1103.13&  & 327.301910 & \textbf{327.075247} & 5/20  & 18.52  \\
MDPII3\_5000  &  & \textbf{324.813456} & \textbf{324.801590} & 10/20 & 955.81  &  & 324.810826 & 324.790223 & 4/20  & 27.82  \\
MDPII4\_5000  &  & \textbf{322.237586} & \textbf{322.197276} & 5/20  & 664.10  &  & 322.212289 & 322.126605 & 4/20  & 338.02 \\
MDPII5\_5000  &  & \textbf{322.491211} & \textbf{322.380726} & 7/20  & 1014.90 &  & 322.420806 & 322.301249 & 5/20  & 105.12 \\
MDPII6\_5000  &  & 322.950488 & \textbf{322.703887} & 4/20  & 352.88  &  & 322.950488 & 322.615227 & 5/20  & 212.55 \\
MDPII7\_5000  &  & 322.850438 & \textbf{322.793125} & 10/20 & 714.31  &  & 322.850438 & 322.778396 & 8/20  & 756.14 \\
MDPII8\_5000  &  & \textbf{323.112120}  & \textbf{323.053268} & 11/20 & 879.48 &  & 323.033840 & 322.873156 & 5/20  & 161.94 \\
MDPII9\_5000  &  & \textbf{323.543775} & \textbf{323.339842} & 7/20  & 569.73  &  & 323.522709 & 323.278556 & 3/20  & 148.94 \\
MDPII10\_5000 &  & 324.519908 & \textbf{324.414458} & 15/20 & 752.95  &  & 324.519908 & 324.294790 & 10/20 & 753.14 \\
\hline
\#Better       &  & \centering{11}    & \centering{19}&   &        &  & \centering{2}    & \centering{1}   &  &   & \\
\#Equal        &  & \centering{7}     & \centering{0} &   &        &  & \centering{7}    & \centering{0}   &  &   & \\
\#Worse        &  & \centering{2}     & \centering{1} &   &        &  & \centering{11}   & \centering{19}  &  &   & \\
\textit{p-value}        &  & \centering{$1.26e-2$}& \centering{$5.699e-5$} &   &       &  &     &    &  &   & \\
\hline
\end{tabular}
\end{tiny}
\end{table}
\renewcommand{\baselinestretch}{1.0}\large\normalsize

Table \ref{results_LS_vs_MA} discloses that for the 20 instances with $n=3000$ the MAMMDP algorithm performs slightly better than the MTS algorithm, but the differences is small. However, for the 20 larger instances with $n=5000$, the MAMMDP algorithm significantly outperforms the the MTS algorithm. First, compared with the MTS algorithm, the MAMMDP algorithm obtains better and worse results in terms of the best objective value on 11 and 2 instances respectively. Second, in terms of average objective value, the MAMMDP algorithm yields better results on 19 out of 20 instances. In addition, from the Friedman test, one observes that the obtained \textit{p-values} are $1.26e-2$ (<0.05) and $5.699e-5$ (<0.05) respectively for the best and average objective values, implying there exists a significant difference between these two methods. These outcomes demonstrate that the memetic framework is particularly useful to solve large and difficult instances.

\section{Conclusions }
\label{Concluesion}

In this paper, we propose the first population-based memetic algorithm (MAMMDP) for solving the NP-hard max-mean dispersion problem (MaxMeanDP). MAMMDP integrates an effective tabu search procedure and a random crossover operator while adopting an original scheme for parent selection. The computational results on a large number of 200 benchmark instances show that the proposed algorithm is very competitive compared with the state-of-the-art algorithms in the literature. Specifically, it improves or matches the previous best known results for all tested instances with $n\le 1000$ with an average computing time of less than 14 seconds and a success rate of 100\%, with only one exception. In particular, we found new and improved best results for 59 out of the 60 most challenging instances. We also show computational results on 40 large instances with 3000 or 5000 elements which can serve as reference lower bounds for evaluating new MaxMeanDP algorithms.

The investigations of several important ingredients confirm that both the underlying tabu search procedure and the crossover operator of the proposed algorithm contribute to the high performance of the proposed algorithm. It is shown that the population-based memetic framework is particularly suitable to solve large and difficult problem instances.

The proposed algorithm could be adapted to the weighted version of the max-mean dispersion problem with several small modifications. Some ideas of the proposed algorithm could be applied to other other binary optimization problems (including some dispersion problems) where no constraint is imposed on the number of variables taking the value of one.

\section*{Acknowledgments}
The work is partially supported by the LigeRo project (2009-2014) and a post-doc grant for X.J. Lai from the Region of Pays de la Loire (France) and the PGMO (2014-0024H) project from the Jacques Hadamard Mathematical Foundation.

%\bibliographystyle{}

%\appendix
%\section{Appendix}
%\section*{Appendix}
\renewcommand{\baselinestretch}{0.65}\scriptsize
\begin{table}[!htp]\centering
\caption{\textbf{Appendix}: Computational results of the proposed MAMMDP algorithm on the set of 100 small instances with $n\le150$. MAMMDP is run 20 times to solver each instance, each run being limited to 10 seconds. The previous best known results ($f_{pre}$) are from: http://www.optsicom.es/edp/}\label{Results_MA_Very_Small}
\begin{tiny}
\begin{tabular}{p{1.8cm}p{0.8cm}p{0.8cm}p{0.01cm}p{1.5cm}p{1.5cm}p{0.8cm}p{0.8cm}p{0.01cm}}
\hline
  &   &   &   & \multicolumn{4}{c}{Memetic Algorithm}  & \\
\cline{5-8}
Instance  &  n  & \centering{$f_{pre}$} &   & $f_{best}$ & $f_{avg}$ & SR & $t(s)$ & \\
\hline
MDPI1\_20    & 20  & 13.88   &  & 13.880000     & 13.880000    & 20/20 & 0.02 &  \\
MDPI2\_20    & 20  & 13.608  &  & 13.608000     & 13.608000    & 20/20 & 0.02 &  \\
MDPI3\_20    & 20  & 11.7957 &  & 11.795714     & 11.795714    & 20/20 & 0.02 &  \\
MDPI4\_20    & 20  & 17.54   &  & 17.540000     & 17.540000    & 20/20 & 0.02 &  \\
MDPI5\_20    & 20  & 16.0063 &  & 16.006250     & 16.006250    & 20/20 & 0.02 &  \\
MDPI6\_20    & 20  & 14.6064 &  & 14.606364     & 14.606364    & 20/20 & 0.02 &  \\
MDPI7\_20    & 20  & 14.8822 &  & 14.882222     & 14.882222    & 20/20 & 0.02 &  \\
MDPI8\_20    & 20  & 14.4614 &  & 14.461429     & 14.461429    & 20/20 & 0.02 &  \\
MDPI9\_20    & 20  & 14.035  &  & 14.035000     & 14.035000    & 20/20 & 0.02 &  \\
MDPI10\_20   & 20  & 13.4433 &  & 13.443333     & 13.443333    & 20/20 & 0.02 &  \\
MDPII1\_20   & 20  & 18.855  &  & 18.855000     & 18.855000    & 20/20 & 0.02 &  \\
MDPII2\_20   & 20  & 17.83   &  & 17.830000     & 17.830000    & 20/20 & 0.02 &  \\
MDPII3\_20   & 20  & 18.11   &  & 18.110000     & 18.110000    & 20/20 & 0.02 &  \\
MDPII4\_20   & 20  & 17.842  &  & 17.842000     & 17.842000    & 20/20 & 0.02 &  \\
MDPII5\_20   & 20  & 16.344  &  & 16.344000     & 16.344000    & 20/20 & 0.02 &  \\
MDPII6\_20   & 20  & 17.61   &  & 17.610000     & 17.610000    & 20/20 & 0.02 &  \\
MDPII7\_20   & 20  & 18.9383 &  & 18.938333     & 18.938333    & 20/20 & 0.02 &  \\
MDPII8\_20   & 20  & 21.88   &  & 21.880000     & 21.880000    & 20/20 & 0.02 &  \\
MDPII9\_20   & 20  & 19.785  &  & 19.785000     & 19.785000    & 20/20 & 0.02 &  \\
MDPII10\_20  & 20  & 22.599  &  & 22.599000     & 22.599000    & 20/20 & 0.02 &  \\
MDPI1\_25    & 25  & 17.2708 &  & 17.270833 & 17.270833 & 20/20 & 0.03 &  \\
MDPI2\_25    & 25  & 15.1214 &  & 15.121429 & 15.121429 & 20/20 & 0.02 &  \\
MDPI3\_25    & 25  & 14.1817 &  & 14.181667 & 14.181667 & 20/20 & 0.03 &  \\
MDPI4\_25    & 25  & 19.8567 &  & 19.856667 & 19.856667 & 20/20 & 0.03 &  \\
MDPI5\_25    & 25  & 17.5371 &  & 17.537143 & 17.537143 & 20/20 & 0.02 &  \\
MDPI6\_25    & 25  & 17.9667 &  & 17.966667 & 17.966667 & 20/20 & 0.03 &  \\
MDPI7\_25    & 25  & 16.207  &  & 16.207000 & 16.207000 & 20/20 & 0.03 &  \\
MDPI8\_25    & 25  & 18.1367 &  & 18.136667 & 18.136667 & 20/20 & 0.03 &  \\
MDPI9\_25    & 25  & 17.4778 &  & 17.477778 & 17.477778 & 20/20 & 0.03 &  \\
MDPI10\_25   & 25  & 19.4592 &  & 19.459167 & 19.459167 & 20/20 & 0.03 &  \\
MDPII1\_25   & 25  & 21.81   &  & 21.810000 & 21.810000 & 20/20 & 0.03 &  \\
MDPII2\_25   & 25  & 22.185  &  & 22.185000 & 22.185000 & 20/20 & 0.03 &  \\
MDPII3\_25   & 25  & 23.5644 &  & 23.564444 & 23.564444 & 20/20 & 0.03 &  \\
MDPII4\_25   & 25  & 19.74   &  & 19.740000 & 19.740000 & 20/20 & 0.03 &  \\
%MDPII5\_25   & 25  & 20.6444 &  & \textbf{20.790000} & \textbf{20.790000} & 20/20 & 0.03 &  \\
MDPII6\_25   & 25  & 20.1744 &  & 20.174444 & 20.174444 & 20/20 & 0.03 &  \\
MDPII7\_25   & 25  & 19.947  &  & 19.947000 & 19.947000 & 20/20 & 0.03 &  \\
MDPII8\_25   & 25  & 23.921  &  & 23.921000 & 23.921000 & 20/20 & 0.03 &  \\
MDPII9\_25   & 25  & 25.016  &  & 25.016000 & 25.016000 & 20/20 & 0.03 &  \\
MDPII10\_25  & 25  & 23.575  &  & 23.575000 & 23.575000 & 20/20 & 0.03 &  \\
MDPI1\_30    & 30  & 19.861  &  & 19.861250 & 19.861250 & 20/20 & 0.03 &  \\
MDPI2\_30    & 30  & 18.813  &  & 18.813333 & 18.813333 & 20/20 & 0.03 &  \\
MDPI3\_30    & 30  & 15.249  &  & 15.248889 & 15.248889 & 20/20 & 0.03 &  \\
MDPI4\_30    & 30  & 22.717  &  & 22.717333 & 22.717333 & 20/20 & 0.03 &  \\
MDPI5\_30    & 30  & 17.237  &  & 17.236667 & 17.236667 & 20/20 & 0.03 &  \\
MDPI6\_30    & 30  & 18.375  &  & 18.375455 & 18.375455 & 20/20 & 0.03 &  \\
MDPI7\_30    & 30  & 15.293  &  & 15.292500 & 15.292500 & 20/20 & 0.03 &  \\
MDPI8\_30    & 30  & 19.247  &  & 19.247273 & 19.247273 & 20/20 & 0.04 &  \\
MDPI9\_30    & 30  & 22.004  &  & 22.004286 & 22.004286 & 20/20 & 0.03 &  \\
MDPI10\_30   & 30  & 18.698  &  & 18.698462 & 18.698462 & 20/20 & 0.03 &  \\
MDPII1\_30   & 30  & 22.2721 &  & 22.272143 & 22.272143 & 20/20 & 0.03 &  \\
MDPII2\_30   & 30  & 26.9138 &  & 26.913846 & 26.913846 & 20/20 & 0.03 &  \\
MDPII3\_30   & 30  & 21.8973 &  & 21.897273 & 21.897273 & 20/20 & 0.04 &  \\
MDPII4\_30   & 30  & 20.5375 &  & 20.537500 & 20.537500 & 20/20 & 0.03 &  \\
MDPII5\_30   & 30  & 22.79   &  & 22.790000 & 22.790000 & 20/20 & 0.03 &  \\
MDPII6\_30   & 30  & 20.351  &  & 20.351000 & 20.351000 & 20/20 & 0.03 &  \\
MDPII7\_30   & 30  & 27.655  &  & 27.655000 & 27.655000 & 20/20 & 0.03 &  \\
MDPII8\_30   & 30  & 26.8842 &  & 26.884167 & 26.884167 & 20/20 & 0.03 &  \\
MDPII9\_30   & 30  & 24.1767 &  & 24.176667 & 24.176667 & 20/20 & 0.03 &  \\
MDPII10\_30  & 30  & 24.8    &  & 24.800000 & 24.800000 & 20/20 & 0.03 &  \\
MDPI1\_35    & 35  & 19.1833 &  & 19.183333 & 19.183333 & 20/20 & 0.04 &  \\
MDPI2\_35    & 35  & 17.168  &  & 17.168000 & 17.168000 & 20/20 & 0.04 &  \\
MDPI3\_35    & 35  & 17.0746 &  & 17.074615 & 17.074615 & 20/20 & 0.04 &  \\
MDPI4\_35    & 35  & 23.35   &  & 23.350000 & 23.350000 & 20/20 & 0.04 &  \\
MDPI5\_35    & 35  & 19.0177 &  & 19.017692 & 19.017692 & 20/20 & 0.04 &  \\
MDPI6\_35    & 35  & 19.445  &  & 19.445000 & 19.445000 & 20/20 & 0.04 &  \\
MDPI7\_35    & 35  & 19.4971 &  & 19.497143 & 19.497143 & 20/20 & 0.04 &  \\
MDPI8\_35    & 35  & 21.2307 &  & 21.230667 & 21.230667 & 20/20 & 0.04 &  \\
MDPI9\_35    & 35  & 20.98   &  & 20.980000 & 20.980000 & 20/20 & 0.04 &  \\
MDPI10\_35   & 35  & 16.9378 &  & 16.937778 & 16.937778 & 20/20 & 0.04 &  \\
MDPII1\_35   & 35  & 25.968  &  & 25.967500 & 25.967500 & 20/20 & 0.04 &  \\
MDPII2\_35   & 35  & 26.135  &  & 26.135455 & 26.135455 & 20/20 & 0.04 &  \\
MDPII3\_35   & 35  & 24.159  &  & 24.159231 & 24.159231 & 20/20 & 0.04 &  \\
MDPII4\_35   & 35  & 24.415  &  & 24.415000 & 24.415000 & 20/20 & 0.04 &  \\
MDPII5\_35   & 35  & 23.857  &  & 23.857500 & 23.857500 & 20/20 & 0.04 &  \\
MDPII6\_35   & 35  & 24.673  &  & 24.673077 & 24.673077 & 20/20 & 0.04 &  \\
MDPII7\_35   & 35  & 29.394  &  & 29.393846 & 29.393846 & 20/20 & 0.04 &  \\
MDPII8\_35   & 35  & 25.297  &  & \emph{25.217273} & \emph{25.217273} & 20/20 & 0.04 &  \\
MDPII9\_35   & 35  & 27.435  &  & 27.435000 & 27.435000 & 20/20 & 0.04 &  \\
MDPII10\_35  & 35  & 25.712  &  & 25.712500 & 25.712500 & 20/20 & 0.04 &  \\
MDPI1\_150   & 150 & 45.92   &  & 45.920192 & 45.920192 & 20/20 & 0.17 &  \\
MDPI2\_150   & 150 & 43.39   &  & 43.392381 & 43.392381 & 20/20 & 0.19 &  \\
MDPI3\_150   & 150 & 40.05   &  & 40.046304 & 40.046304 & 20/20 & 0.17 &  \\
MDPI4\_150   & 150 & 44.04   &  & 44.044138 & 44.044138 & 20/20 & 0.17 &  \\
MDPI5\_150   & 150 & 42.48   &  & 42.479388 & 42.479388 & 20/20 & 0.17 &  \\
MDPI6\_150   & 150 & 43.72   &  & 43.722955 & 43.722955 & 20/20 & 0.17 &  \\
MDPI7\_150   & 150 & 46.08   &  & 46.077308 & 46.077308 & 20/20 & 0.16 &  \\
MDPI8\_150   & 150 & 42.45   &  & 42.451346 & 42.451346 & 20/20 & 0.18 &  \\
MDPI9\_150   & 150 & 42.48   &  & 42.479767 & 42.479767 & 20/20 & 0.18 &  \\
MDPI10\_150  & 150 & 41.8    &  & 41.797805 & 41.797805 & 20/20 & 0.16 &  \\
MDPII1\_150  & 150 & 57.48   &  & 57.484000 & 57.484000 & 20/20 & 0.16 &  \\
MDPII2\_150  & 150 & 57.82   &  & 57.820652 & 57.820652 & 20/20 & 0.17 &  \\
MDPII3\_150  & 150 & 58.42   &  & 58.421818 & 58.421818 & 20/20 & 0.17 &  \\
MDPII4\_150  & 150 & 57.38   &  & 57.381064 & 57.381064 & 20/20 & 0.17 &  \\
MDPII5\_150  & 150 & 54.23   &  & 54.228571 & 54.228571 & 20/20 & 0.17 &  \\
MDPII6\_150  & 150 & 56.44   &  & 56.442653 & 56.442653 & 20/20 & 0.16 &  \\
MDPII7\_150  & 150 & 58.89   &  & 58.889167 & 58.889167 & 20/20 & 0.17 &  \\
MDPII8\_150  & 150 & 57.97   &  & 57.965370 & 57.965370 & 20/20 & 0.17 &  \\
MDPII9\_150  & 150 & 58.3    &  & 58.302619 & 58.302619 & 20/20 & 0.17 &  \\
MDPII10\_150 & 150 & 57.18   &  & 57.175122 & 57.175122 & 20/20 & 0.16 &  \\
\hline
\end{tabular}
\end{tiny}
\end{table}
\renewcommand{\baselinestretch}{1.0}\large\normalsize

\end{document}